\documentclass{article} 
\usepackage{iclr2026_conference,times}

\usepackage{amsmath,amsfonts,bm}

\def\eqref#1{equation~\ref{#1}}

\def\1{\bm{1}}

\DeclareMathAlphabet{\mathsfit}{\encodingdefault}{\sfdefault}{m}{sl}
\SetMathAlphabet{\mathsfit}{bold}{\encodingdefault}{\sfdefault}{bx}{n}

\usepackage{hyperref}
\usepackage{url}
\usepackage{graphicx}
\usepackage{graphicx}
\usepackage{caption}
\usepackage{subcaption}
\usepackage[most]{tcolorbox}
\tcbuselibrary{skins,breakable,listings}
\usepackage{inconsolata}

\newtcblisting{promptbox}{
  listing only,
  title      = {LLM Selector Prompt for Molecular Selection},
  colback    = black!1,
  colframe   = black!40,
  boxrule    = 0.5pt,
  arc        = 2pt,
  left=2mm, right=2mm, top=1mm, bottom=1mm,
  fonttitle  = \bfseries,
  listing options={basicstyle=\footnotesize\ttfamily,breaklines=true},
  breakable,
}

\title{LLM-Guided Retrieval for Prediction of Molecular Perturbation Responses} 

\author{Betty Xiong \thanks{Work completed while employed at Genentech.}
\thanks{Correspondence to \texttt{xiongb\{at\}stanford.edu} and \texttt{maleki.sepideh\{at\}gene.com.}}\\
Department of Biomedical Data Science \\
Stanford University\\
Stanford, CA 94305, USA \\
\texttt{\{xiongb\}@stanford.edu} \\
\AND
Jan-Christian Huetter, Gabriele Scalia, Tommaso Biancalani \& Sepideh Maleki \\
Biology Research \& AI Development, Genentech \\
DNA Way, South San Francisco, CA 94080, USA \\
\texttt{\{huetter.janchristian-klaus, scalia.gabriele, biancalani.tommaso, maleki.sepideh\}@gene.com} \\
}

\iclrfinalcopy 
\begin{document}

\maketitle

\begin{abstract}

Predicting transcriptomic responses to small-molecule perturbations across cell lines is central to drug discovery, but exhaustive profiling of drug--cell combinations is infeasible. We frame molecular perturbation prediction as \emph{retrieve-and-aggregate}: approximate an unmeasured drug's response in a cell line by aggregating measured responses of a small set of biologically related compounds. We propose \textit{LLM-Guided Retrieval (LGR)}, where a large language model (LLM) ranks candidate neighbor drugs (restricted to those profiled in the target cell line); after which a fixed mean aggregator combines their observed expression deltas to form the prediction. We evaluate on the Tahoe-100M single-cell perturbation atlas under unseen-drug, unseen-cell-line, and open-world regimes. LGR consistently improves over drug mean, ChemCPA, and chemistry-based kNN baselines, with the strongest gains for unseen cell-line generalization, where it achieves higher correlation and lower error than mean baselines. Across settings, LGR improves directional (sign) accuracy of gene regulation, indicating better recovery of biologically meaningful perturbation effects even when magnitude-based metrics are similar. These results suggest that retrieval quality--rather than predictor complexity--is a key driver of zero-shot molecular perturbation prediction, and that LLMs can provide a useful biological prior when used as constrained retrieval modules.

\end{abstract}

\section{Introduction}

Systematically predicting how small molecules perturb gene expression in specific cellular contexts is a central capability for drug discovery and functional genomics. Large transcriptomics perturbation resources have demonstrated that expression signatures can connect compounds, pathways, and disease states, but they still cover only a small fraction of the combinatorial space of (cell line, drug) conditions \citep{cmap,l1000}. Recent single-cell chemical perturbation atlases further expand this landscape and enable cell-line--resolved effect estimation, yet comprehensive profiling across cell lines and compounds remains elusive \citep{sciplex,tahoe}.

A common approach is to learn a supervised mapping from compound and basal expression to transcriptomic outcomes, leveraging single-cell perturbation data to generalize to unseen conditions. Methods based on latent-variable modeling and compositional generalization can perform well for predicting perturbation responses under certain regimes \citep{scgen,cpa,chemcpa}. However, these models can be sensitive to distribution shift, and their gains can depend strongly on how similar test-time compounds and contexts are to those observed during training.

In parallel, benchmarks and analyses have found that simple baselines remain surprisingly strong in perturbation modeling, and that progress often hinges on selecting informative analogs or priors rather than increasing predictor complexity \citep{linearbaselines,perturbench,op3}. This motivates a complementary view: for many zero-shot or few-shot settings, the bottleneck may be retrieval---identifying biologically appropriate neighbors---more than learning a complex decoder.

We therefore frame molecular perturbation prediction as \emph{retrieve-and-aggregate}. Our selector–aggregator framing is inspired by recent work on genetic perturbations that uses an LLM (\texttt{GPT-5.2}) to select informative neighbors for a $k$ nearest neighbors (kNN)-style predictor of unseen gene perturbation effects \citep{langpert}. We adapt this idea to small-molecule perturbations under a cell-line–restricted candidate pool. Given a cell-line and drug pair with an unobserved transcriptomic effect, we retrieve a small neighborhood of biologically related compounds that were profiled in the same cell line, then aggregate their observed expression deltas to form the prediction. Within this framing, the key modeling question is how to construct a high-quality, cell-line--specific neighborhood without access to the query transcriptome.

We introduce \textbf{LLM-Guided Retrieval (LGR)}, which uses a large language model as a constrained selector over a candidate pool of compounds measured in the target cell line. By leveraging the LLM's mechanistic and pathway-level knowledge while enforcing a closed candidate set, LGR provides a lightweight biological prior that can be paired with a transparent mean aggregator, and show that out-of-distribution (OOD) perturbation prediction hinges more strongly on selecting the correct priors than on model expressivity. We evaluate LGR on the Tahoe-100M single-cell perturbation atlas \citep{tahoe} under unseen-drug, unseen-cell-line, and open-world regimes, and find that LGR consistently improves over drug mean and chemistry-based kNN baselines, with the strongest gains for unseen cell-line generalization. These results suggest that retrieval quality is a key driver of zero-shot perturbation prediction and that LLMs can be effective as constrained retrieval modules for biological reasoning.

This work makes the following contributions:
\begin{itemize}
    \item We formulate molecular perturbation prediction as a retrieve-and-aggregate problem,
    providing a simple and interpretable alternative to end-to-end supervised models.
    \item We introduce \textit{LLM-Guided Retrieval (LGR)}, a framework in which a large language
    model is used solely as a selector to identify biologically relevant perturbations, while all numerical computation is performed by a fixed downstream aggregation operator.
    \item Through empirical evaluation on large-scale single-cell perturbation data, we show that LGR achieves its strongest gains when generalizing to unseen cell lines and remains competitive with strong cell-mean baselines in open-world settings.
    
\end{itemize}

\section{Related Work}

\paragraph{Transcriptomic perturbation datasets.}
Transcriptomic perturbation resources have long supported a ``signature'' view of perturbations, where gene expression profiles can be compared to connect compounds, pathways, and disease states. The Connectivity Map and subsequent LINCS/L1000 efforts operationalized this paradigm at scale, enabling retrieval-style analog reasoning over perturbation signatures \citep{cmap,l1000}. More recent single-cell perturbation atlases extend this idea to cell-line--resolved responses, providing substantially richer contexts for modeling and evaluation. For example, sci-Plex profiles transcriptional responses across multiple cancer cell lines and a diverse compound panel \citep{sciplex}, while Tahoe-100M scales to a large single-cell perturbation atlas with broad treatment coverage across many cancer models \citep{tahoe}. Related community efforts such as the Arc Institute Virtual Cell Challenge further emphasize prediction of cellular responses to perturbations as a core benchmark task \citep{virtual}. Together, these resources motivate evaluation protocols that explicitly test generalization across drugs and cellular contexts.

\paragraph{Perturbation response prediction.}
A large body of work studies predictive models of gene expression responses to perturbations from observed (cell line, perturbation) pairs. Early and widely used approaches learn mappings between control and perturbed states (e.g., \textsc{scGen} \citep{scgen}), or represent perturbations in latent spaces designed for compositional generalization (\textsc{CPA} \citep{cpa} and \textsc{chemCPA} \citep{chemcpa}).
Other methods, including variational frameworks such as \textsc{scVIDR} \citep{scvidr}, further explore flexible latent-variable models for response prediction. More recently, benchmarks have stressed OOD evaluation--notably unseen-drug and unseen-cell-line-settings--and shown that reported gains can depend strongly on split design and leakage controls \citep{op3, perturbench}. In particular, analyses have found that simple baselines can remain surprisingly competitive, sometimes matching or exceeding more complex predictors under certain evaluation regimes \citep{linearbaselines}. These observations motivate our emphasis on controlled unseen-drug and unseen-cell-line regimes, and our focus on isolating the role of neighborhood construction rather than relying on increasingly expressive decoders. We also note that many existing perturbation models (including compositional latent approaches) implicitly rely on additive or approximately linear assumptions in representation space; our formulation makes this assumption explicit and measurable.

\paragraph{LLMs as selectors and biological priors for retrieval.}
Beyond supervised predictors, a complementary line of work treats model inference as selection or retrieval: the core challenge becomes identifying informative analogs, examples, or actions, with downstream prediction performed by a simple rule. LLMs have been increasingly used in this decision module role, where they select candidates or tools rather than directly generating high-dimensional numeric outputs. In biomedicine, LLMs encode pharmacological and mechanistic knowledge important for relating drugs by targets, pathways, and functional similarity \citep{llmencode, biogpt}. Closest to our approach, \citet{langpert} uses an LLM to select informative neighbors for predicting the effects of unseen \emph{genetic} perturbations with a kNN-style estimator. We adapt this selector--aggregator idea to \emph{small-molecule} perturbations under a cell-line--restricted candidate pool: the LLM is constrained to rank only compounds profiled in the target cell line, and prediction is performed by a fixed mean aggregator over retrieved expression deltas. This design isolates the contribution of retrieval quality while keeping the downstream predictor transparent.

\begin{figure}[t]
    \centering
    \includegraphics[width=0.8\linewidth]{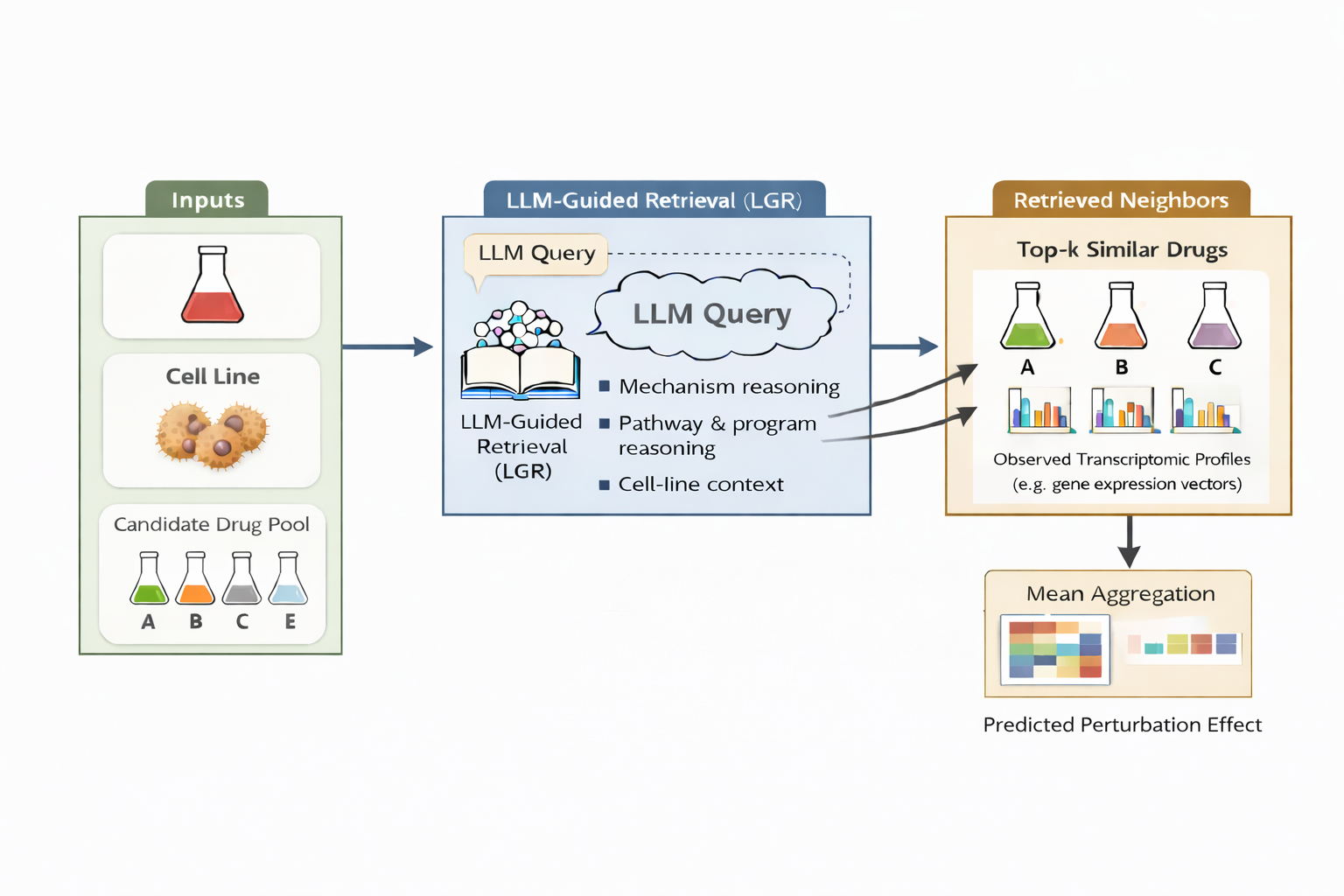}
    \caption{
    \textbf{Overview of LLM-Guided Retrieval (LGR).}
    Given a query drug and a target cell type, LGR uses a large language model (LLM) as a
    constrained retrieval module to identify pharmacologically similar drugs from a restricted
    candidate pool. 
    The transcriptomic response of the query drug is then estimated by aggregating (mean pooling)
    the observed responses of the top-$k$ retrieved drugs in the same cell type.
    }
    \label{fig:lgr_overview}
\end{figure}

\vspace{-2mm}
\section{Methodology}
\label{sec:methods}

We study the problem of predicting the transcriptomic effect of a small-molecule perturbation on a specific cell line.
Let $\mathcal{D}$ denote a set of drugs and $\mathcal{C}$ denote a set of cell lines. For a drug $d \in \mathcal{D}$ and a cell line $c \in \mathcal{C}$, we denote by $\Delta_{c,d}\in\mathbb{R}^G$ the transcriptomic effect vector (control-subtracted perturbation effect) over $G$ genes. Our objective is to estimate \(\Delta_{c,d}\) for query pairs \((c,d)\) where \(\Delta_{c,d}\) is unobserved.


\paragraph{Similarity-based effect estimation.}

For a query drug $d$ and cell line $c$, we define a neighborhood $\mathcal{N}_k(c,d) \subseteq \mathcal{D} \setminus \{d\}$ consisting of the $k$ drugs whose transcriptomic effects are expected to be most similar to that of $d$ in cell line $c$. We estimate the transcriptomic effect of $d$ on $c$ by aggregating the observed effects of its
neighbors.

\begin{equation}
\widehat{\Delta}_{c,d}
\;=\;
\mathrm{Agg}\big(\{\Delta_{c,d'}: d'\in\mathcal{N}_k(c,d)\}\big).
\end{equation}

In this work, $\mathrm{Agg}(\cdot)$ is the uniform mean (Section~\ref{sec:aggregation}).
Under this formulation, the primary challenge is identifying an informative, cell-line--specific
neighborhood $\mathcal{N}_k(c,d)$ without using the query's transcriptome.

\paragraph{Candidate pool.}For each cell line $c$, we define the candidate pool
\begin{equation}
\mathcal{P}_c \;=\; \{ d' \in \mathcal{D} : \Delta_{c,d'} \text{ is measured and } d' \text{ is non-control} \}.
\end{equation}
For a query drug $d$ in cell line $c$, retrieval is performed over $\mathcal{P}_c \setminus \{d\}$ to avoid trivial self-matches. Non-control specifies i.e., excluding vehicle (e.g., DMSO) controls. This pool restriction ensures that all retrieved neighbors have observed transcriptomic effects in the target cell line, so the downstream aggregation is well-defined.

\subsection{LLM-Guided Retrieval}
\label{sec:lgr}

We propose \textbf{LLM-Guided Retrieval (LGR)}, which uses a large language model as a \emph{selector} to rank candidate neighbor drugs from a cell-line-specific pool (Figure~\ref{fig:lgr_overview}).

\paragraph{Selector input and output.}
Given (i) a query drug name $d$, (ii) a cell line $c$, and (iii) a list of candidate drug names $\mathcal{P}_c \setminus \{d\}$, the LLM is prompted to return a ranked list of the top-$k$ candidates most likely to induce similar transcriptomic effects in cell line $c$. The prompt explicitly instructs the model to: (i) select only from the provided candidate list, ii) consider mechanistic and pathway-level relationships (rather than chemical similarity alone), and (iii) provide a brief justification for each choice. Any out-of-pool or invalid outputs were discarded at runtime, and final neighbors are validated against canonical compound identifiers. Prompt template details can be found in Appendix~\ref{app:prompt}.

\paragraph{Mechanism and program inference.}

At runtime, LGR associates each drug with a set of fine-grained mechanism tags
(e.g., EGFR inhibition, CDK4/6 inhibition, PARP inhibition) inferred from internal pharmacological knowledge.
These mechanisms are mapped to higher-level transcriptomic programs, including RTK/MAPK signaling,
PI3K--AKT--mTOR signaling, cell cycle regulation, DNA damage and replication stress, and epigenetic
or proteostasis stress.

\paragraph{Parsing and validation.}
We parse the LLM output to extract the ranked drug names and apply lightweight normalization (e.g., lowercasing and whitespace stripping). Each predicted neighbor is then validated against the candidate list $\mathcal{P}_c$; out-of-pool items are discarded and duplicates are removed while preserving rank order. The resulting realized neighbor set has size $k_{\mathrm{used}} \le k$:
\begin{equation}
\mathcal{N}_k(c,d) \;=\; \{d'_1,\ldots,d'_{k_{\mathrm{used}}}\} \subseteq \mathcal{P}_c \setminus \{d\}.
\end{equation}
When $k_{\mathrm{used}} < k$, we aggregate over the valid retrieved items only. We record $k_{\mathrm{used}}$ for each query to characterize selector coverage and its impact on variance. 

\paragraph{Aggregation operator.}
\label{sec:aggregation}
Given a validated neighbor set $\mathcal{N}_k(c,d)$, our main predictor uses uniform mean aggregation:
\begin{equation}
\widehat{\Delta}_{c,d}
\;=\;
\frac{1}{|\mathcal{N}_k(c,d)|}
\sum_{d' \in \mathcal{N}_k(c,d)} \Delta_{c,d'}.
\end{equation}
This choice is deliberately simple and transparent: it isolates the contribution of retrieval quality from that of a learned downstream predictor.

\paragraph{Reproducibility via caching.}
We cache validated neighbor lists $\mathcal{N}_k(c,d)$ per query, including the raw
LLM output and postprocessing outcomes (e.g., filtered items and $k_{\mathrm{used}}$), to ensure
reproducibility and enable ablations.

\section{Experiments}

\paragraph{Dataset and preprocessing.}
\label{sec:data}
We evaluate on the Tahoe-100M single-cell perturbation atlas \citep{tahoe}. We log-normalize counts, compute pseudobulk means per (cell line, drug) condition, and compute control-subtracted deltas using matched vehicle controls. We restrict evaluation to a subset of cell lines with sufficient coverage across perturbations. Detailed information on controls, splits and candidate pools can be found in Appendix~\ref{app:data}.

\paragraph{LLM implementation details.}
LLM-Guided Retrieval (LGR) uses a large language model as a constrained selector for inducing
drug similarity at runtime. All experiments in this work use a \texttt{GPT-5}–class model with deterministic
decoding (temperature set to zero). The model is queried using a fixed prompt template and a
restricted candidate pool, and its outputs are parsed and filtered deterministically as described in Section~\ref{sec:methods}.

\subsection{Evaluation Regimes}
\label{sec:eval_regimes}
We evaluate \textit{LGR} under two complementary settings designed to target different aspects of generalization: 




\paragraph{Closed-world evaluation.}
In the closed-world regime, the candidate pool available to all methods is restricted to drugs observed during training. This setting ensures a leakage-free comparison and places LGR and learning-based baselines under equivalent information access. Within this regime, we consider two tasks:
(i) \emph{unseen drug generalization}, where test drugs are held out across all cell lines, and
(ii) \emph{unseen cell-line generalization}, where all perturbations from a subset of cell lines are
held out during training. These two tasks isolate chemical generalization from cellular-context
generalization.

\paragraph{Open-world evaluation.}
In the open-world regime, the LLM selector is allowed to rank a query drug against the full dataset. This regime reflects realistic use cases for knowledge-driven systems, where prior biological knowledge is available but curated training sets
are not. Results in this setting are reported separately and are not directly compared to supervised
models.

Across all evaluation regimes, the prediction target is the control-subtracted gene expression delta $\Delta_{c,d}$ (Section~\ref{sec:methods}). Unless otherwise specified, the retrieve-and-aggregate predictor uses $k=10$ neighbors.


\subsection{Baselines}
\label{sec:baselines}

To contextualize the performance of our model, we compare against a set of standard and strong baselines commonly used in perturbation prediction and drug--cell response modeling. All baselines are evaluated using the same train--test splits and metrics as the proposed method.
Let $\Delta_{c,d} \in \mathbb{R}^G$ denote the true gene expression change (delta) for cell $c$ under drug $d$, where $G$ is the number of genes.

\paragraph{Cell mean.}
The \textit{cell mean} baseline predicts the average perturbation response of a cell line across all training drugs observed in that cell line:
\begin{equation}
\widehat{\Delta}_{c,d} = \mu_c, \quad
\mu_c = \mathbb{E}_{d':(c,d')\in\mathcal{S}_{\text{train}}}[\Delta_{c,d'}].
\end{equation}
If a test cell line $c$ is not observed during training (e.g., unseen-cell-line regime), we fall back to the global mean $\mu = \mathbb{E}_{(c',d')\in\mathcal{S}_{\text{train}}}[\Delta_{c',d'}]$.

\paragraph{Drug mean.}
The \textit{drug mean} baseline predicts the average effect of a drug across all training cell lines in which it is observed:
\begin{equation}
\widehat{\Delta}_{c,d} = \mu_d, \quad
\mu_d = \mathbb{E}_{c':(c',d)\in\mathcal{S}_{\text{train}}}[\Delta_{c',d}].
\end{equation}
This baseline is cell-agnostic (it predicts the same vector for all $c$) and captures drug-specific transcriptional signatures that generalize across cell lines.

\paragraph{PCA + ridge regression (PCA+RR).}
\label{sec:pca_ridge}

We implement a scalable linear baseline that predicts deltas in a low-dimensional latent space, by combining principal component analysis (PCA) with ridge regression. We fit PCA on training deltas $\{\Delta_{c,d} : (c,d)\in\mathcal{S}_{\text{train}}\}$ and retain the top $K$ components, yielding a projection matrix
$\mathbf{P}\in\mathbb{R}^{G\times K}$. Each training delta is embedded as
\begin{equation}
\mathbf{z}_{c,d} = \mathbf{P}^\top \Delta_{c,d} \in \mathbb{R}^{K}.
\end{equation}

We represent each drug $d$ by a Morgan fingerprint $\mathbf{x}_d$ and train a ridge regressor to predict latent coordinates from drug features:
\begin{equation}
\widehat{\mathbf{z}}_{c,d} = \mathbf{W}\mathbf{x}_d, \qquad
\mathbf{W} = \arg\min_{\mathbf{W}} \sum_{(c,d)\in\mathcal{S}_{\text{train}}}
\left\|\mathbf{z}_{c,d} - \mathbf{W}\mathbf{x}_d\right\|_2^2 + \lambda \|\mathbf{W}\|_F^2.
\end{equation}
Finally, we map back to gene space via the inverse PCA transform:
\begin{equation}
\widehat{\Delta}_{c,d} = \mathbf{P}\widehat{\mathbf{z}}_{c,d}.
\end{equation}

\paragraph{Chemistry kNN.}
To exploit chemical similarity, we implement a kNN baseline in drug fingerprint embedding space. Each drug $d$ is represented by a Morgan fingerprint $\mathbf{x}_d$ derived from its SMILES representation.
For a query drug $d$, we retrieve the $k$ most similar \textit{training} drugs under cosine similarity:
\begin{equation}
\mathcal{N}^{\text{chem}}_k(d) = \operatorname{TopK}_{d'\in\mathcal{D}_{\text{train}}\setminus\{d\}}
\ \cos(\mathbf{x}_{d}, \mathbf{x}_{d'}).
\end{equation}
We then predict by averaging the corresponding drug-mean profiles:
\begin{equation}
\widehat{\Delta}_{c,d} = \frac{1}{k} \sum_{d' \in \mathcal{N}^{\text{chem}}_k(d)} \mu_{d'}.
\end{equation}
This baseline generalizes drug effects based on chemical similarity and does not rely on observing the test drug during training. It is cell-agnostic because it averages drug-level profiles $\mu_{d'}$ rather than cell-specific deltas.

\paragraph{chemCPA.}
We include ChemCPA baseline~\cite{chemcpa} to assess whether learned cell-specific representations combined with molecular features can explain perturbation responses in our setting. The model represents each drug using a fixed Morgan fingerprint derived from its SMILES representation. Since ChemCPA uses fixed cell-line representation, we could not evaluate this model on the unseen cell line generalization task.

\subsection{Evaluation Metrics}
We report Pearson correlation ($r$), cosine similarity, mean absolute error (MAE), mean squared error (MSE), sign accuracy, and coefficient of determination ($R^2$) between predicted and true deltas. Metrics are computed on $\Delta_{c,d}$ in the log-normalized, control-subtracted space. We select the top 2000 HVGs and evaluate all metrics on that gene subset.

\paragraph{Sign accuracy.}

Sign accuracy measures the directional consistency between predicted and true gene expression changes. For a given perturbation $(c,d)$ and gene $g$, let $\hat{\Delta}_{c,d}^{(g)}$ and $\Delta_{c,d}^{(g)}$ denote the predicted and true gene expression deltas, respectively. The sign accuracy for a single perturbation is defined as the fraction of genes for which the predicted and true deltas have the same sign:
\[
\mathrm{SignAcc}_{c,d} = \frac{1}{G} \sum_{g=1}^{G} \mathbb{I}\!\left[\operatorname{sign}\!\left(\hat{\Delta}_{c,d}^{(g)}\right) = \operatorname{sign}\!\left(\Delta_{c,d}^{(g)}\right)\right],
\]
where $\mathbb{I}[\cdot]$ denotes the indicator function and $G$ is the number of genes.

We report sign accuracy averaged over all evaluated perturbations. This metric captures whether a model correctly predicts the direction of up- or down-regulation for each gene, independent of the magnitude of the predicted effect.

\subsection{Results}

\subsubsection{Closed-World Evaluation}

\begin{figure}[t]
  \centering

  \begin{subfigure}[t]{0.32\linewidth}\centering
    \includegraphics[width=\linewidth]{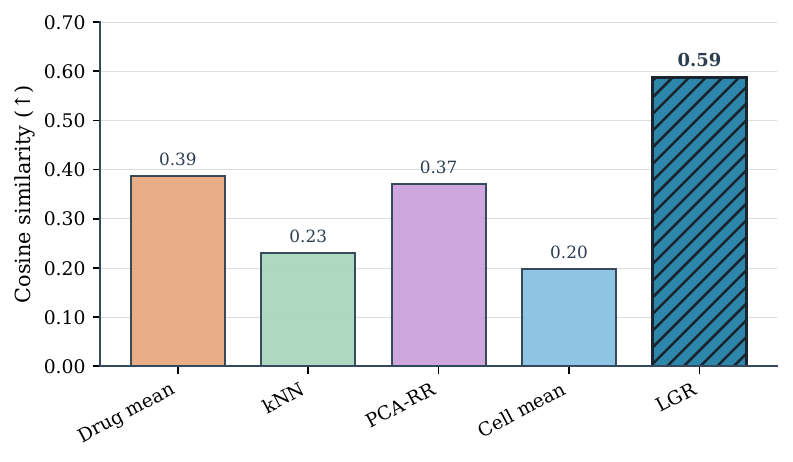}
    \caption{}\label{fig:grid:cosine}
  \end{subfigure}\hfill
  \begin{subfigure}[t]{0.32\linewidth}\centering
    \includegraphics[width=\linewidth]{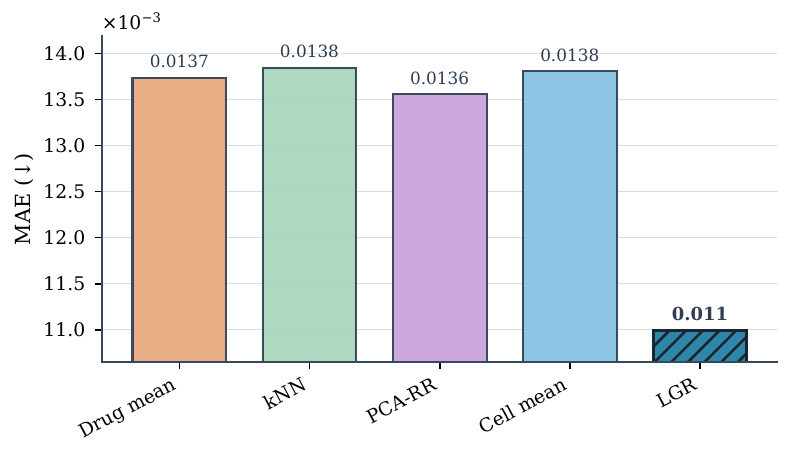}
    \caption{}\label{fig:grid:mae}
  \end{subfigure}\hfill
  \begin{subfigure}[t]{0.32\linewidth}\centering
    \includegraphics[width=\linewidth]{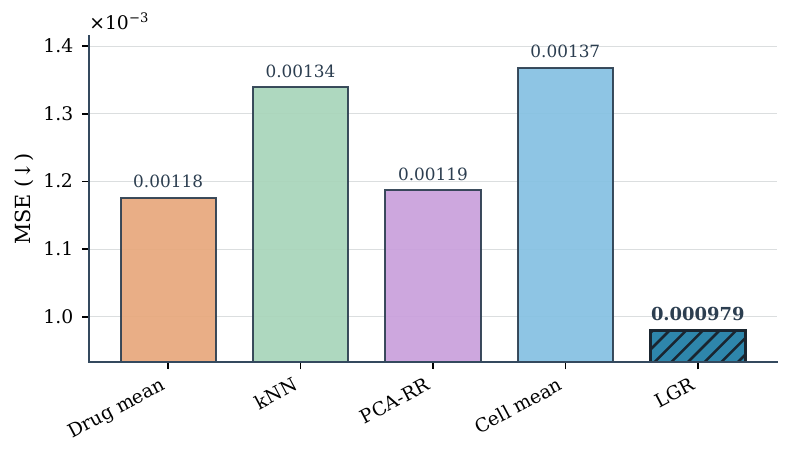}
    \caption{}\label{fig:grid:mse}
  \end{subfigure}


  \begin{subfigure}[t]{0.32\linewidth}\centering
    \includegraphics[width=\linewidth]{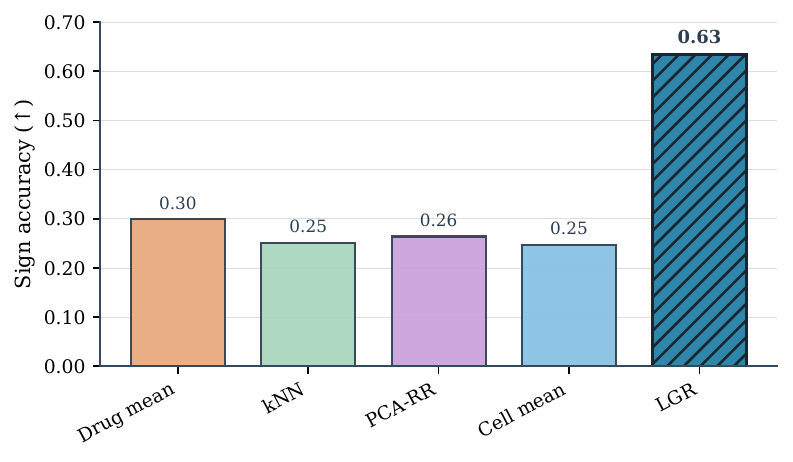}
    \caption{}\label{fig:grid:pds}
  \end{subfigure}\hfill
  \begin{subfigure}[t]{0.32\linewidth}\centering
    \includegraphics[width=\linewidth]{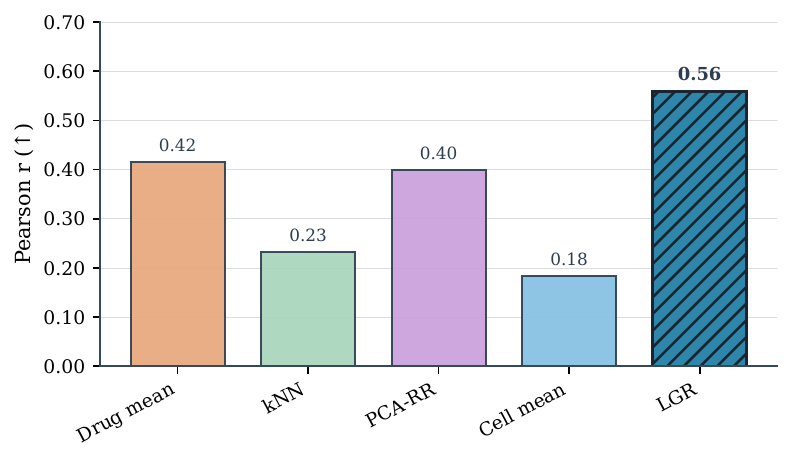}
    \caption{}\label{fig:grid:r}
  \end{subfigure}\hfill
  \begin{subfigure}[t]{0.32\linewidth}\centering
    \includegraphics[width=\linewidth]{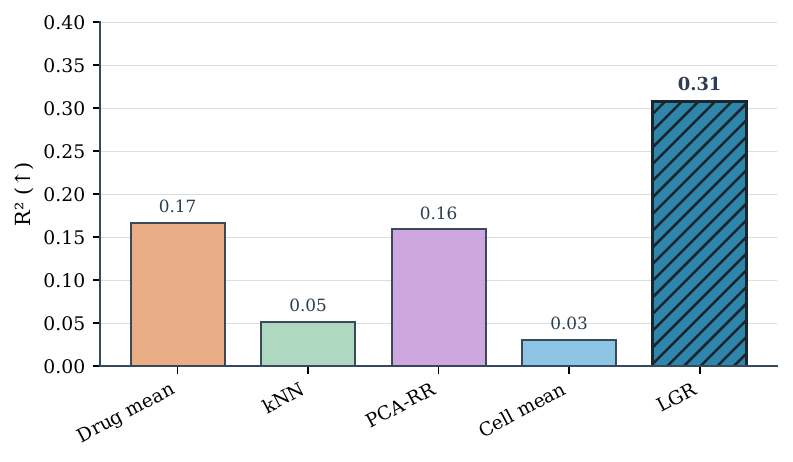}
    \caption{}\label{fig:grid:r2}
  \end{subfigure}

  \caption{Unseen Cell-Line Generalization. Evaluation metrics are (a) cosine similarity, (b) MAE, (c) MSE, (d) sign accuracy, (e) Pearson correlation, and (f) $R^2$ of linear regression.}
  \label{fig:unseen_cell}
\end{figure}

\paragraph{Unseen-cell-line generalization}
Figure~\ref{fig:unseen_cell} evaluates generalization to held-out cell lines, where supervised baselines cannot use any cell-specific training deltas. The performance gain can be attributed to zero-shot learning via literature transfer, i.e. due to the LLM's pre-training on literature about cell lines. LGR achieves best overall performance, improving Pearson correlation by $>0.15$, roughly doubling $R^2$, and reducing MAE/MSE relative to cell-mean and drug-mean baselines. LGR also achieves the highest sign accuracy, consistent with retrieving perturbations that transfer conserved pathway-level programs across cellular contexts.

\begin{figure}[t]
  \centering

  \begin{subfigure}[t]{0.32\linewidth}\centering
    \includegraphics[width=\linewidth]{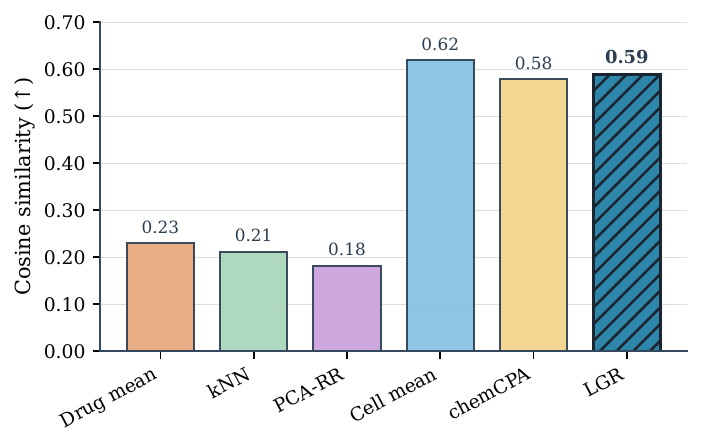}
    \caption{}\label{fig:grid:cosine}
  \end{subfigure}\hfill
  \begin{subfigure}[t]{0.32\linewidth}\centering
    \includegraphics[width=\linewidth]{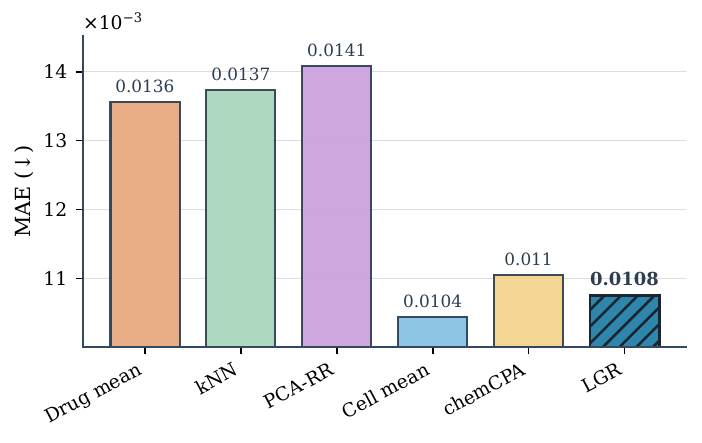}
    \caption{}\label{fig:grid:mae}
  \end{subfigure}\hfill
  \begin{subfigure}[t]{0.32\linewidth}\centering
    \includegraphics[width=\linewidth]{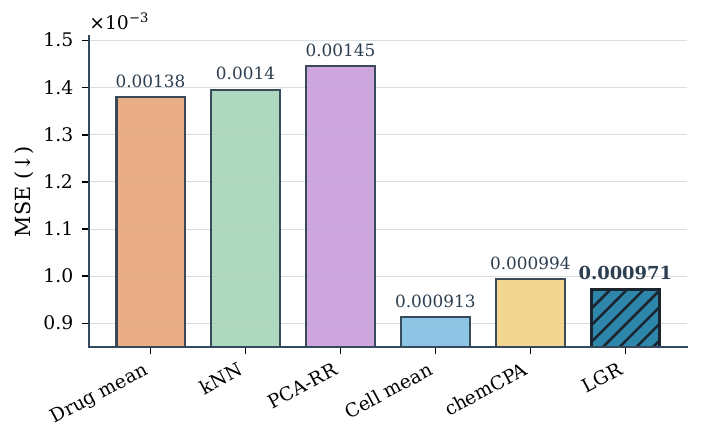}
    \caption{}\label{fig:grid:mse}
  \end{subfigure}


  \begin{subfigure}[t]{0.32\linewidth}\centering
    \includegraphics[width=\linewidth]{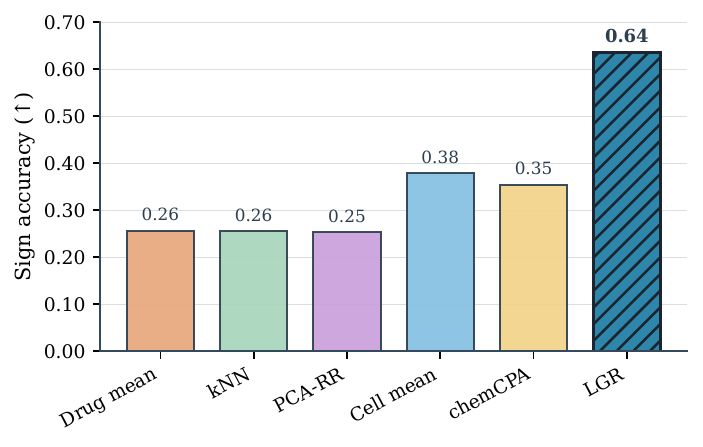}
    \caption{}\label{fig:grid:pds}
  \end{subfigure}\hfill
  \begin{subfigure}[t]{0.32\linewidth}\centering
    \includegraphics[width=\linewidth]{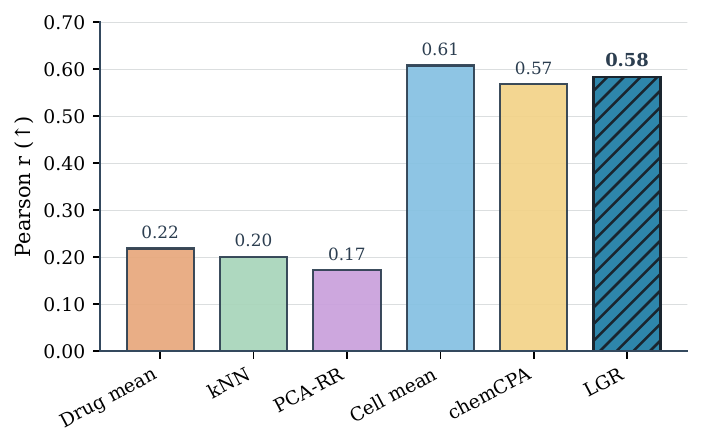}
    \caption{}\label{fig:grid:r}
  \end{subfigure}\hfill
  \begin{subfigure}[t]{0.32\linewidth}\centering
    \includegraphics[width=\linewidth]{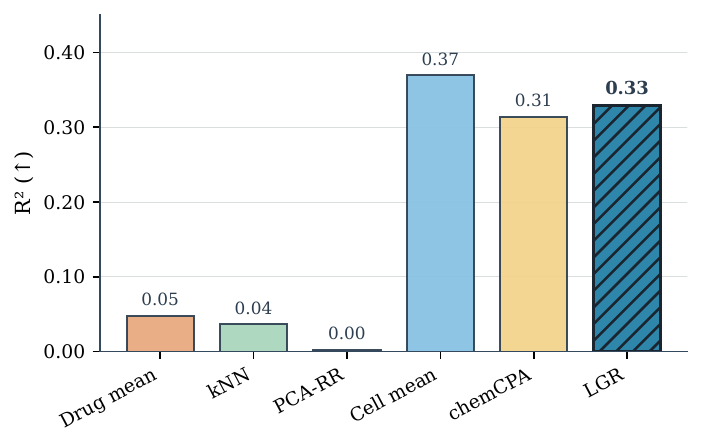}
    \caption{}\label{fig:grid:r2}
  \end{subfigure}

  \caption{Unseen Drug Generalization. Evaluation metrics are (a) cosine similarity, (b) MAE, (c) MSE, (d) sign accuracy, (e) Pearson correlation, and (f) $R^2$ of linear regression.}
  \label{fig:unseen_drug}
\end{figure}

\paragraph{Unseen-drug generalization}
Figure~\ref{fig:unseen_drug} reports performance when test drugs are held out across all cell lines. LGR is competitive with the strongest baselines on correlation and error metrics, and consistently outperforms chemistry kNN and PCA+RR. While the cell-mean baseline is strong in this regime, LGR achieves the best sign accuracy, with nearly a two-fold improvement over drug-mean and kNN.




This gain in sign accuracy indicates that LGR more reliably captures the \emph{direction} of gene
regulation induced by unseen drugs, even when absolute effect sizes are difficult to calibrate.

\subsubsection{Open-World Evaluation}
\begin{figure}[t]
  \centering

  \begin{subfigure}[t]{0.32\linewidth}\centering
    \includegraphics[width=\linewidth]{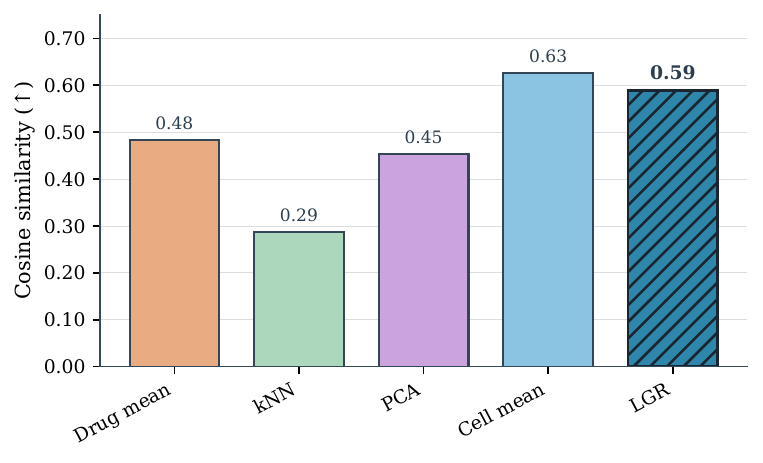}
    \caption{}\label{fig:grid:cosine}
  \end{subfigure}\hfill
  \begin{subfigure}[t]{0.32\linewidth}\centering
    \includegraphics[width=\linewidth]{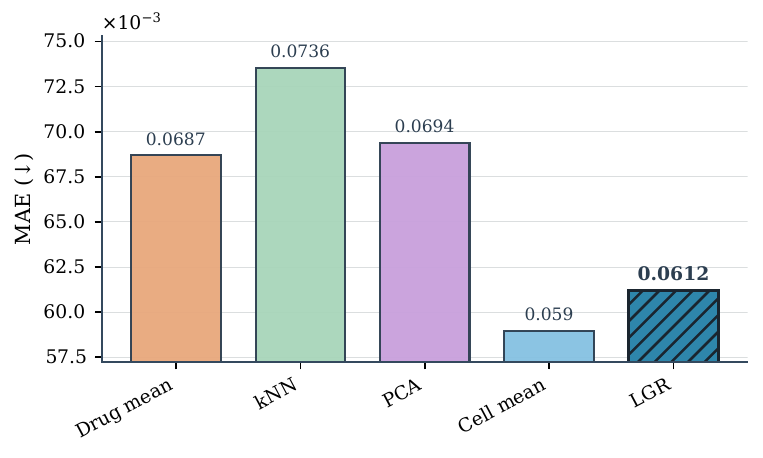}
    \caption{}\label{fig:grid:mae}
  \end{subfigure}\hfill
  \begin{subfigure}[t]{0.32\linewidth}\centering
    \includegraphics[width=\linewidth]{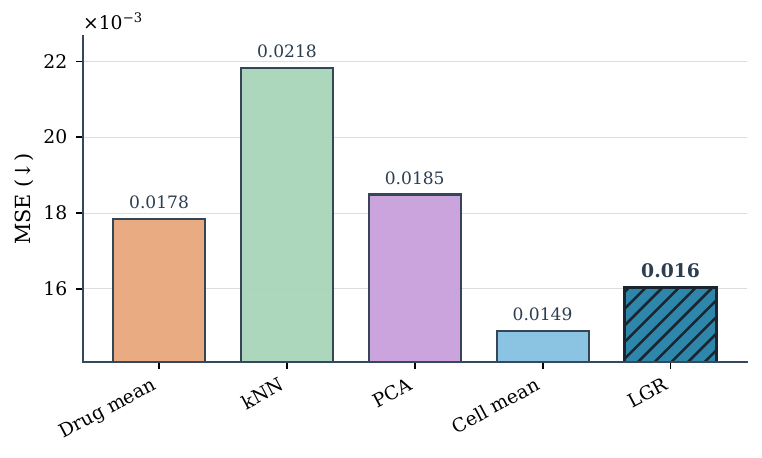}
    \caption{}\label{fig:grid:mse}
  \end{subfigure}


  \begin{subfigure}[t]{0.32\linewidth}\centering
    \includegraphics[width=\linewidth]{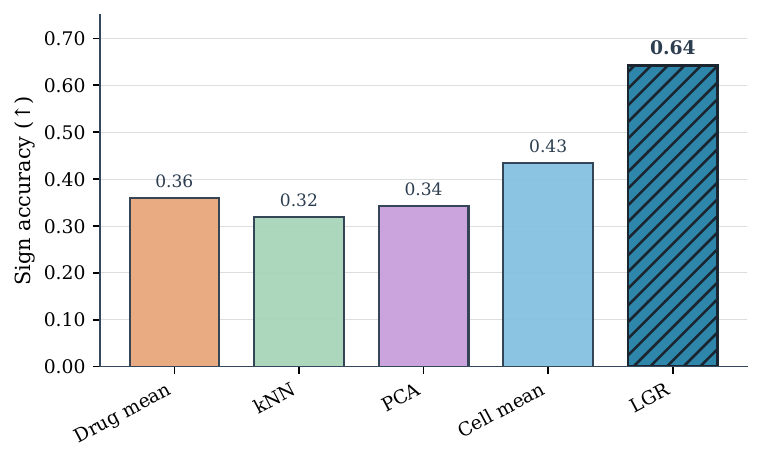}
    \caption{}\label{fig:grid:pds}
  \end{subfigure}\hfill
  \begin{subfigure}[t]{0.32\linewidth}\centering
    \includegraphics[width=\linewidth]{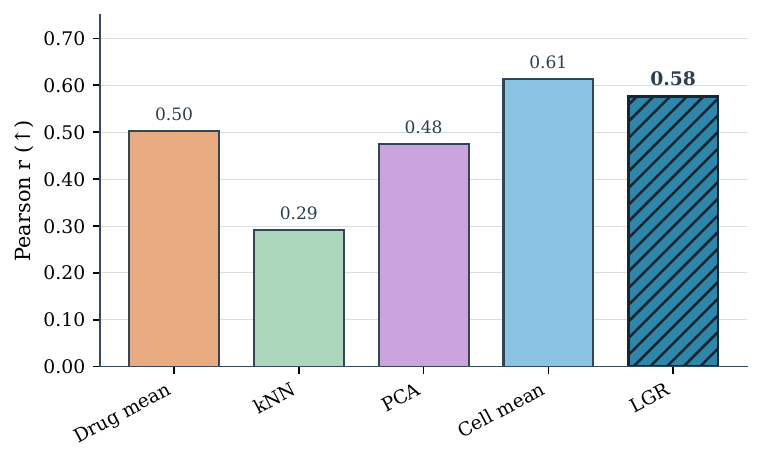}
    \caption{}\label{fig:grid:r}
  \end{subfigure}\hfill
  \begin{subfigure}[t]{0.32\linewidth}\centering
    \includegraphics[width=\linewidth]{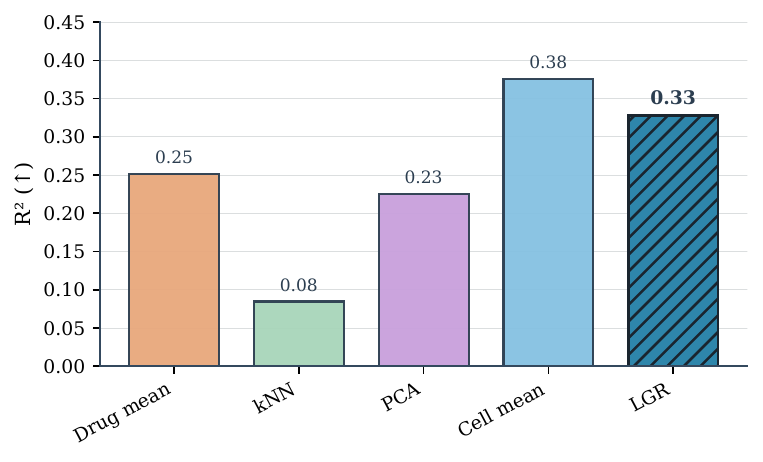}
    \caption{}\label{fig:grid:r2}
  \end{subfigure}

  \caption{Open-World Evaluation. Evaluation metrics are (a) cosine similarity, (b) MAE, (c) MSE, (d) sign accuracy, (e) Pearson correlation, and (f) $R^2$ of linear regression.}
  \label{fig:all}
\end{figure}

Figure~\ref{fig:all} shows open-world performance when retrieval is allowed over all profiled compounds in the target cell line. LGR remains competitive with the cell-mean baseline on magnitude-based metrics while outperforming all other baselines. In particular, LGR improves sign accuracy by $>20$ percentage points over cell mean, indicating more reliable recovery of the qualitative direction of regulation.




These results highlight a key strength of LGR: while simple averaging baselines can match overall response magnitudes, LGR more reliably recovers the qualitative structure of transcriptional responses. This directional accuracy is particularly important for downstream biological interpretation, such as pathway enrichment and mechanism-of-action analysis, where the sign of
gene regulation often matters more than precise effect size.

\subsection{Discussion}

Across evaluation regimes, our results suggest that retrieval quality is often as important as
predictor complexity. Holding the aggregation operator fixed
(Section~\ref{sec:aggregation}), LLM-Guided Retrieval (LGR) improves over simple similarity
baselines and achieves its strongest gains in the unseen cell-line regime, where generalization
across cellular context is required. In this setting, LGR outperforms mean-based baselines across
correlation, error, and directional metrics, indicating that biologically informed neighborhood
selection is particularly valuable when cell-specific statistics are unavailable.

A consistent pattern across experiments is that improvements in directional (sign) accuracy are
larger than gains in magnitude-based metrics. We hypothesize that this reflects a calibration
effect: uniform mean aggregation tends to shrink predictions toward shared transcriptional
programs (or toward the cell mean when neighborhoods are noisy), limiting improvements in
magnitude-sensitive metrics. In contrast, LGR retrieves neighbors that better preserve which
transcriptional programs are activated or repressed in the target context, leading to improved
directional accuracy even when effect sizes remain difficult to match. More expressive
aggregation schemes, such as rank- or confidence-weighted averaging, may help address this
calibration gap.

We observe three primary failure modes of LGR: (i) out-of-pool or aliasing errors that reduce the
effective number of retrieved neighbors ($k_{\mathrm{used}}$), (ii) uncertainty on less well-studied
compounds that yields generic or weakly informative matches, and (iii) instability under uniform
averaging when few neighbors remain. These observations motivate a simple hybrid strategy in
which LGR is applied when coverage is high, and deterministic chemistry-based retrieval is used as
a fallback otherwise. Next steps would include biological analysis of retrieved neighbors, and optimization of the number of neighbors. Overall, our results suggest that large language models are most effective
when used as constrained retrieval modules, while the downstream predictor can remain simple,
transparent, and non-parametric.

\subsubsection*{Acknowledgments}
BX is supported by Australian-American Fulbright Commission Future Scholarship.

\bibliography{iclr2026_conference}
\bibliographystyle{iclr2026_conference}
\clearpage
\appendix

\section{LLM Selector Prompt, Parsing, and Filtering}
\label{app:prompt}
\paragraph{Prompting with a restricted candidate pool.}

\begin{promptbox}

Candidate Molecules
You are given the following list of candidate molecules:
{candidate_molecules}

Task
Given an anchor molecule and a cell line, your task is to identify the top k molecules from the candidate list that are most likely to induce similar transcriptomic effects in the specified cell line.

Your reasoning should explicitly consider:
* known or inferred mechanisms of action,
* affected signaling pathways or transcriptional programs,
* pathway adjacency or mechanistic overlap,
* whether these programs are expected to be active or dominant in the given cell line.

Do not rely on chemical similarity alone.
Do not introduce molecules outside the provided candidate list.
Your goal is to select molecules whose biological perturbation programs are most likely to resemble those of the anchor molecule in the given cellular context.

Output Format
Return exactly k molecules, ranked from most to least similar.
Use the following format:

<Final Answer>
1. MOLECULE 1: One-sentence biological justification.
2. MOLECULE 2: One-sentence biological justification.
...
</Final Answer>

Query
Anchor molecule: {molecule}
Cell line: {cell_type}

\end{promptbox}

\paragraph{Parsing.}
We parse the \texttt{<Final Answer>} block and extract the molecule string before the first colon on each numbered line.
We apply light normalization (lowercasing; stripping whitespace; collapsing repeated spaces; removing trailing punctuation).

\paragraph{Canonicalization and candidate validation.}
Each parsed name is checked against the candidate list \(\mathcal{P}_c\). If the model outputs an out-of-pool molecule or an invalid name, we \textit{ignore it}. If fewer than $k$ valid unique molecules remain, we aggregate over the remaining valid ones only. Thus, the realized neighbor count satisfies \(k_{\text{used}} \le k\).
Duplicates (after normalization) are removed while preserving order.

\paragraph{Realized \(k_{\text{used}}\) and coverage.}
We record \(k_{\text{used}}\) for each query and method. Low \(k_{\text{used}}\) is a primary source of LLM variance: when few valid neighbors remain after filtering, the mean aggregation becomes noisy and can collapse toward near-zero slopes.

\paragraph{Caching neighbors.}
For reproducibility, neighbor lists are cached per query as JSONL:
\begin{quote}\small
\texttt{\{target\_cell\_type, target\_molecule, neighbors:[\{molecule, score\}]\}}
\end{quote}
where \texttt{score} is optional (e.g., Tanimoto similarity). LLM neighbors omit scores by default.

\section{Data Processing, Splits, and Candidate Pools}
\label{app:data}

\paragraph{Dataset.}
We use the Tahoe-100M dataset, a mega-scale single-cell perturbation atlas with 100M+ cells, \(\sim\)60k experiments, 50 cancer models, and 1,100+ treatments. We subsample 100 cell lines within the full dataset.
We compute pseudo-bulk expression profiles per observed (cell line, molecule) and log-normalize counts.

\paragraph{Controls and deltas.}
For each cell line \(c\), the control mean profile \(\mathbf{u}_c\) is computed by averaging control wells.
For each observed pair \((c,d)\), the perturbation delta is \(\Delta_{c,d} = \mathbf{y}^{\mathrm{raw}}_{c,d} - \mathbf{u}_c\).

\paragraph{Splits.}
Train/validation/test query pairs are specified by Arrow files containing \((\texttt{cell\_type}, \texttt{molecule})\).
All retrieval evaluation is conducted on the test set pairs.

\paragraph{Candidate pools.}
For each cell line \(c\), we define the candidate pool \(\mathcal{P}_c\) as all non-control molecules measured in \(c\).
For each query \((c,d)\), candidates are \(\mathcal{P}_c \setminus \{d\}\).

\section{Reproducibility Notes}
\label{app:repro}

\paragraph{Implementation.}
We implement data handling with AnnData; neighbor lists are cached as JSONL per query; aggregation and metrics are vectorized in NumPy.
We log run configuration (seed, \(k\), aggregation mode, metric settings, and split identifiers).

\end{document}